\pdfoutput=1

\documentclass[11pt]{article}

\usepackage[]{EMNLP2023}

\usepackage{times}
\usepackage{latexsym}

\usepackage{graphicx}
\usepackage{amssymb}
\usepackage{amsmath}    
\usepackage{cleveref}

\usepackage[T1]{fontenc}

\usepackage[utf8]{inputenc}

\usepackage{microtype}

\usepackage{inconsolata}

\usepackage{comment}

\usepackage{booktabs,multirow}

\usepackage{verbatim} %
\usepackage{graphicx} %
\title{Identifying and Adapting Transformer-Components Responsible for Gender Bias in an English Language Model}

\author{Abhijith Chintam$^{12}$, Rahel Beloch$^1$\\ $^1$Master AI, University of Amsterdam, $^2$Pegasystems, Amsterdam, The Netherlands\\\texttt{archintam@gmail.com, mail@rahelbeloch.de}
\AND Willem Zuidema, Michael Hanna\Thanks{ Shared senior authorship.}, Oskar van der Wal\footnotemark[1]\\ Institute for Logic, Language \& Computation, University of Amsterdam\\\texttt{\{w.h.zuidema, m.w.hanna, o.d.vanderwal\}@uva.nl}}

\begin{document}

\maketitle

\begin{abstract}
Language models (LMs) exhibit and amplify many types of undesirable biases learned from the training data, including gender bias. However, we lack tools for effectively and efficiently changing this behavior without hurting general language modeling performance.
In this paper, we study three methods for identifying causal relations between LM components and particular output: causal mediation analysis, automated circuit discovery and our novel, efficient method called DiffMask+ based on differential masking.
We apply the methods to GPT-2 small and the problem of gender bias, and use the discovered sets of components to perform parameter-efficient fine-tuning for bias mitigation.
Our results show significant overlap in the identified components (despite huge differences in the computational requirements of the methods) as well as
success in mitigating gender bias, with less damage to general language modeling compared to full model fine-tuning. 
However, our work also underscores the difficulty of defining and measuring bias, and the sensitivity of causal discovery procedures to dataset choice. 
We hope our work can contribute to more attention for dataset development, and lead to more effective mitigation strategies for other types of bias.

\end{abstract}

\section{Introduction}

Modern neural language models exhibit social biases, such as biases based on gender, religion, ethnicity and other \emph{protected attributes}. These biases may lead to real harms when used in down-stream applications \cite[e.g.][]{hovy-spruit-2016-social,Weidinger2021EthicalAS}. Detecting and mitigating biases in language models has therefore become an important area of research. 

Early detection methods relied on lists of words to measure associations with e.g., specific genders \cite[e.g.][]{Caliskan_Bryson_Narayanan_2017}. Most current detection methods work with curated sets of sentence pairs or triplets, and measure differences in sentence probabilities or anaphora resolution probabilities 
\cite[e.g.][]{May_Wang_Bordia_Bowman_Rudinger_2019,Nadeem_Bethke_Reddy_2021,nangia-etal-2020-crows,Basta_Costa-jussà_Casas_2019}. 
Proposed mitigation strategies include targeted changes to the training data \citep[e.g., CDA;][]{lu2020gender}, training procedure \citep[e.g., adversarial learning;][]{zhang2018mitigating}, model parameters \citep[e.g., INLP;][]{ravfogel-etal-2020-null}, or language generation procedure \citep[e.g., ``self-debiasing'';][]{schick-etal-2021-self}. %

Despite this work, we still lack a proper understanding of how to best measure biases (how do we guarantee the representativeness for real-world harm of a set of sentence pairs, or of a linguistic phenomenon such as anaphora resolution?), how biases are implemented in the language model internals (is there a unified locus, or is, e.g., gender bias the aggregate effect of many independent model decisions?), and what techniques are effective at reducing undesirable downstream behavior (e.g., is data curation more or less effective than filtering output? Is intervening in the model internals feasible?).
Empirically, success in detecting and mitigating biases depends on many factors, including the choice of embeddings, training regimes, data sets and model choices 
\cite{blodgett2020language,blodgett-etal-2021-stereotyping,talat2022you,Delobelle_Tokpo_Calders_Berendt_2022,barrett-etal-2019-adversarial,van-der-wal-etal-2022-birth}.

The ``black-box'' nature of LMs makes it difficult to identify and interpret how bias manifests and propagates in them, especially relying solely on correlational methods. The starting point for the current paper is the intuition that if, instead, it were possible to find \emph{causal} relationships between the model's internal representations and its downstream bias, we could more effectively measure and intervene on these undesirable behaviors.

We therefore turn to a recent series of papers on interpretability methods that focus on causal discovery. 
In \Cref{sec:relatedwork} we discuss three such methods, of which we adapt one (DiffMask) for our needs in 
\Cref{sec:representation-cma-circuits}. Our new method is more efficient than other causal methods, which is especially relevant when applied to large language models (LLMs). In \Cref{sec:representation-cma-circuits} we also report results from these three methods when applied to GPT2-small and the problem of gender bias, and find that they discover largely overlapping sets of components, despite huge differences in computation requirements.
In \Cref{sec:intervention} we use the identified components  to adapt GPT-2 small, using parameter-efficient fine-tuning procedures. 
We demonstrate how gender bias in LMs can be reduced with minimal effect to their language modelling performance by making targeted interventions to their components.
However, we also recognize the limitations of operationalizing gender bias as we do, using minimal pairs of contrasting sentences---which simplify gender as a \emph{binary} construct and may not work so well for other languages than English---and call for future research to develop reliable and validated bias measures \citep[see][]{van2023undesirable}.

\section{Related Work}
\label{sec:relatedwork}
Where and how LMs implement output behaviors---from high-level phenomena like gender stereotypes, to lower-level ones like subject-verb agreement---is an active field of study. In providing an overview of related work, we focus on causal methods for locating mechanisms in \cref{sec:locating_mechanisms}, as non-causal methods can yield misleading conclusions \citep{ravichander-etal-2021-probing,elazar-etal-2021-amnesic}. Further, we review previous work on targeted changes to Language models and their behavior in \cref{sec:targeted_changes}

\subsection{Locating Mechanisms in Language Models} \label{sec:locating_mechanisms}

Causal methods study model processing by intervening in (altering) model processing, and observing the changes in model behavior caused by these interventions. They aim to address the shortcomings in observational methods by ensuring a causal link between mechanisms found in model internals, and model behavior.

Many such techniques determine which representations or components are important to model processing by ablating them. Ablations can range from zeroing out neurons \citep{lakretz-etal-2019-emergence,mohebbi-etal-2023-quantifying}, to replacing them with a baseline \citep{Cao_Schlichtkrull_Aziz_Titov_2021, Bau2018IdentifyingAC}, or replacing them with another example's activation \citep{Vig_Gehrmann_Belinkov_Qian_Nevo_Singer_Shieber_2020,Geiger_Lu_Icard_Potts_2021}. All of these techniques return unstructured sets of important components without specifying their interaction.

In recent years, the \emph{circuits} abstraction of transformer models \citep{elhage2021mathematical} has become popular. This framework views transformer models as computational graphs, and aims to find subgraphs responsible for certain tasks. This technique has been used to find circuits for indirect object detection and the greater-than operation in GPT-2 \citep{Wang_Variengien_Conmy_Shlegeris_Steinhardt_2023,hanna2023does}, as well as to study larger models \citep{lieberum2023does}; it has also been automated \citep{Conmy_Mavor-Parker_Lynch_Heimersheim_Garriga-Alonso_2023}. 

Note that although causal methods can provide a higher degree of confidence in localizing mechanisms, they are not foolproof. For example, \citet{Meng_Bau_Andonian_Belinkov_2023} propose causal tracing, a method for locating fact storage in LMs; they then edit GPT-2 XL's factual knowledge by performing edits at relevant locations. However, recent work has showed that although edits may be successful, the localization found by causal tracing is not predictive of edit success \citep{hase2023does}. So, even causal localizations should be assessed thoroughly.

\subsection{Targeted Changes to Language Models and Their Behavior} \label{sec:targeted_changes}
One way to mitigate bias in LMs is to change their parameters or internal representations; however, making large changes can be computationally expensive and have unintended side-effects on model behavior. Past work has studied how to make targeted changes to LMs that avoid these pitfalls.
We only discuss works on intervening in the model's representations and parameter-efficient fine-tuning on curated datasets, but other bias mitigation strategies exist as well \citep[see e.g.,][]{meade-etal-2022-empirical}.

\paragraph{Model Interventions}
One line of research focuses on removing undesirable concepts from a LM's representations directly. Early methods like \emph{hard-debias} based on principal component analysis \cite{bolukbasi2016man} and \emph{iterated null-space projection} \citep[INLP,][]{ravfogel-etal-2020-null} identify and remove linear representations of gender (bias) from embedding spaces; while others make targeted changes to the activations of LMs~\cite{Cao_Schmid_Hupkes_Titov_2021,belrose2023leace} or edit the components directly~\cite{meng2022mass,Meng_Bau_Andonian_Belinkov_2023}.

Altering activations at run-time is one promising way to mitigate (gender) bias in LMs. LEACE \cite{belrose2023leace}, for example, convincingly removes linearly-encoded gender information from activations. Similarly, \citet{Cao_Schmid_Hupkes_Titov_2021} use an approach called \emph{differentiable masking} (DiffMask) to identify small neuron subsets responsible for bias and intervene on them for reducing bias.

However, a downside of these activation-altering methods is that they require an intervention on the activations at each inference step. %
Moreover, it is not obvious which model activations we should run these on; for instance, it is unlikely that we want to remove gender information from every input token.

\paragraph{Parameter-Efficient Fine-tuning}

Another approach that avoids some of the pitfalls of changing the LM's representations directly, is to fine-tune on a carefully constructed dataset. %
Previous work has shown the importance of considering the training data in understanding the biases learned by LMs~\cite[e.g.,][]{zhao-etal-2018-gender,zmigrod2019counterfactual,bordia2019identifying,lu2020gender,bender2021dangers,Sellam2022multiberts,van-der-wal-etal-2022-birth,biderman2023pythia}.
Given this, fine-tuning on curated datasets is a promising strategy for mitigating gender bias in LMs \citep{solaiman2021palms,levy-etal-2021-collecting-large,gira-etal-2022-debiasing,kirtane-anand-2022-mitigating}. Falling within this paradigm is \emph{parameter-efficient} fine-tuning, where only some of the model parameters are updated---this may not only be computationally more efficient, but even yield better results~\citep[][]{lauscher-etal-2021-sustainable-modular,gira-etal-2022-debiasing,Xie_Lukasiewicz_2023}.

Our work is most similar to \citet{gira-etal-2022-debiasing}, who also use parameter-efficient fine-tuning for debiasing GPT-2 small. However, we study the effect of fine-tuning individual attention heads, while they focus on embedding layers, LayerNorm parameters, adding linear input/output transformation parameters, and a combination thereof. Moreover, \citeauthor{gira-etal-2022-debiasing} do not adhere to any specific strategy when selecting the components to fine-tune. In contrast, our method provides a principled approach to identify the components that are causally important for the task at hand and then fine-tune them. 

\citeposs{Xie_Lukasiewicz_2023} work is also related to ours. They verify the effectiveness of parameter-efficient bias mitigation techniques like adapter tuning \citep{Houlsby2019ParameterEfficientTL} and prefix tuning \citep{Li2021PrefixTuningOC} on various types of LMs and biases. These methods introduce extra tuneable parameters instead of directly tuning the model parameters themselves.

Our approach could mitigate gender bias to an extent with minimal degradation in language modelling performance, similar to the results of \citet{Xie_Lukasiewicz_2023} and \citet{gira-etal-2022-debiasing}. However, making a direct comparison is challenging due to differences in evaluation criteria and employed datasets. \citet{gira-etal-2022-debiasing} exclusively assess their method on StereoSet \citep{Nadeem_Bethke_Reddy_2021}, whereas we have evaluated our approach on multiple benchmarks, as discussed in \Cref{metrics}. \citet{Xie_Lukasiewicz_2023} evaluate their fine-tuning methods using similar benchmarks as ours, but they employ the older CrowS-Pairs \citep{nangia-etal-2020-crows} dataset for stereotype score and WikiText2 \citep{Merity2016PointerSM} for perplexity. We use a newer, improved version of CrowS-Pairs \citep{neveol2022french} and the much larger WikiText-103 \citep{Merity2016PointerSM} instead.

\section{Locating Gender Bias}
\label{sec:representation-cma-circuits}
In this section, we investigate the question: where in a given LM is gender bias introduced? We study this in GPT-2 small \citep{radford2019language}, an English-language, auto-regressive pre-trained transformer LM.\footnote{The code for our experiments can be found here: \url{https://github.com/iabhijith/bias-causal-analysis}} Its small size---12 transformer layers, with 12 attention heads and 1 multi-layer perceptron (MLP) each---makes it a good object of close studies like we perform. We seek to identify the subset of the 144 attention heads that introduce gender bias into the last position of GPT-2's input, where GPT-2 produces next-token predictions. We identify these heads in the context of inputs that lead to gender-biased next-tokens from GPT-2.

This study thus focuses on attention heads. Though prior work has emphasized the role of MLPs in gender bias and memorization \citep{Vig_Gehrmann_Belinkov_Qian_Nevo_Singer_Shieber_2020,geva-etal-2022-transformer,Meng_Bau_Andonian_Belinkov_2023}, we argue that attention heads are also an interesting subject of analysis. Unless the final word of the input contains gender information that causes the production of biased next-tokens, this information must be introduced from other positions via attention heads.

To determine where GPT-2 small introduces gender bias into its output, we use three methods: causal mediation analysis (CMA), automated circuit discovery, and our own novel method that combines the first approach with differential masking. We then compare the results of these three methods.

\subsection{Methodology}
\label{discovery_methodology}
All methods we use rely on a core technique as outlined in \citet{Vig_Gehrmann_Belinkov_Qian_Nevo_Singer_Shieber_2020}: swapping model component activations during a forward pass on one input, with activations taken from the model when run on another input which induces an opposite behaviour in the model. For this purpose, we use the \textit{Professions dataset} from \citet{Vig_Gehrmann_Belinkov_Qian_Nevo_Singer_Shieber_2020}, which contains templated sentences designed to elicit gender bias. 
The sentences in the dataset take the form ``The \{profession\} said that''. GPT-2's continuations on these sentences tend to be stereotypical---if the profession is \emph{nurse}, GPT-2 outputs \emph{she}, while if it is \emph{doctor}, GPT-2 outputs \emph{he}. 

For each sentence in the dataset we generate a corresponding counterfactual sentence with the profession word replaced by anti-stereotypical gender-specific word. If the normal sentence's profession is female-stereotyped, its corresponding counterfactual sentence is ``The \emph{man} said that''; for male-stereotyped professions, the counterfactual contains \emph{woman}.  These sentences are designed to maximize the change in model behavior with respect to the predicted pronoun; this makes it easier to identify important components.  The dataset contains sentences generated from 17 templates and 299 professions resulting in 5083 sentences in total. For all methods that follow, we intervene on the last position of the sentence.

\subsubsection{Causal Mediation Analysis} \label{sec:cma}
\citet{Vig_Gehrmann_Belinkov_Qian_Nevo_Singer_Shieber_2020} were the first to use CMA~\citep{pearl2014cma} to locate gender bias in GPT-2; we adopt their methods as a baseline. CMA relies on a simple hypothesis: if a component is important to the model's behavior on a task, swapping its output activation with another will change model behavior. More formally, let $\mathbf{x}$ and $\mathbf{\tilde{x}}$ be normal and counterfactual inputs respectively, and let $i$ be the index of the component (attention head or MLP) under investigation. We first run the model on $\mathbf{x}$, and observe its output distribution $p(y|\mathbf{x})$, Then, we run the model on $\mathbf{\tilde{x}}$ and save $\mathbf{\tilde{h}}_i$, the counterfactual output of component $i$. Then we run the model on $\mathbf{x}$ again, but replace $\mathbf{h}_i$ with $\mathbf{\tilde{h}}_i$ during the forward pass. This yields an altered model output distribution $\tilde{p}(y|\mathbf{x})$. \citet{Vig_Gehrmann_Belinkov_Qian_Nevo_Singer_Shieber_2020} measure how important a component $i$ is to a model behaviour $b$ using Natural Indirect Effect (NIE), the expected proportional difference in model behavior after intervening on component $i$. If $b_{null}$ is the original behaviour of the model and $b_{i, intv}$ is the behaviour of the model after intervening on component $i$, then NIE can be evaluated as shown in \Cref{NIE}:
\begin{align}
    \text{NIE}(i, b) = \mathbb E_{(\mathbf{x}, \mathbf{\tilde{x}}) \in   \mathcal D} \left[ \frac{b_{i, intv}}{b_{null}} - 1 \right] \label{NIE}
\end{align}

\citet{Vig_Gehrmann_Belinkov_Qian_Nevo_Singer_Shieber_2020} use the definition in \cref{eq:bias} to measure biased behaviour in a LM. It is the ratio of the probabilities assigned by the model to an anti-stereotypical continuation as against a stereotypical continuation given a context. In case of Professions dataset \citep{Vig_Gehrmann_Belinkov_Qian_Nevo_Singer_Shieber_2020}, it is the ratio of probability assigned to anti-stereotypical pronoun versus the probability assigned to stereotypical pronoun. 

\begin{align}
    b(\mathbf{x}) &= \frac{p(y = \text{anti-stereo}| \mathbf{x})}{p(y = \text{stereo} | \mathbf{x})} \label{eq:bias}
\end{align}

The aforementioned technique analyzes individual components; \citeauthor{Vig_Gehrmann_Belinkov_Qian_Nevo_Singer_Shieber_2020} propose two methods to gather a \emph{set} of important components. Using the top-$k$ strategy, they evaluate every component, and select the $k$ components that cause the most change in model behavior. Using the $k$-greedy strategy, they evaluate all components, and add the most impactful one. Then, they evaluate each component again, ablating both it \emph{and} their set; they once again add the most impactful component. They repeat the latter step until they have a set of size $k$.

\subsubsection{Circuit Discovery}
The circuits framework, which views models as computational graphs, provides a related technique for identifying mechanisms in LMs. While \citeauthor{Vig_Gehrmann_Belinkov_Qian_Nevo_Singer_Shieber_2020}'s CMA approach generates a component set (nodes) relevant to a task, the circuits approach generates a set of edges, resulting in a detailed subgraph. However, the underlying methodology is similar to CMA: we ablate edges via swaps, and see which edges hurt performance once ablated. Though our fine-tuning techniques only target nodes (not edges), comparing CMA and circuits localisations of bias could still be insightful.

We use \citeposs{Conmy_Mavor-Parker_Lynch_Heimersheim_Garriga-Alonso_2023} automated circuit discovery code (ACDC) to identify model components relevant to (gender) bias. This technique iteratively tests model edges, removing those that can be ablated without changing task performance. We use ACDC on the same professions dataset as CMA, and measure task performance as the difference in probability assigned to stereotypical and non-stereotypical pronoun continuations.

\subsubsection{Differentiable Masking With CMA}
\label{diffmask_cma}
We finally propose our own method for localizing relevant LM components that combines two approaches: \citeposs{Vig_Gehrmann_Belinkov_Qian_Nevo_Singer_Shieber_2020} CMA and \citeposs{Cao_Schlichtkrull_Aziz_Titov_2021} differentiable masking (DiffMask). Our method is motivated by a notable challenge with CMA, namely, how to select the best size-$k$ subset of model components that contributes to bias. \citeauthor{Vig_Gehrmann_Belinkov_Qian_Nevo_Singer_Shieber_2020}'s two strategies for this (top-$k$ and $k$-greedy as discussed in \Cref{sec:cma}) both have downsides. A top-$k$ strategy assumes that components' importance is independent, while a $k$-greedy strategy is expensive, requiring $k$ evaluations of all components' importance. A full sweep of the search space would be combinatorially expensive.

This combinatorial search problem can be reformulated as an optimization problem using a differentiable relaxation \citep{louizos2018learning, bastings-etal-2019-interpretable, Cao_Schlichtkrull_Aziz_Titov_2021, Cao_Schmid_Hupkes_Titov_2021, schlichtkrull2021interpreting}. DiffMask, proposed by \citet{Cao_Schmid_Hupkes_Titov_2021} precisely apply the reformulation to learn an almost-binary differentiable stochastic mask over a model's components, indicating which are important, and which are not. Unimportant components are those whose outputs can be ablated without changing model behavior.  

We adapt DiffMask in two ways, and label our variant DiffMask+. First, instead of using surrogate models that instantiate distribution per input, we directly learn a distribution for the stochastic mask. This change is crucial because it helps us identify a single, generalizable set of components responsible for bias in the language model across the entire dataset, which is essential for downstream fine-tuning.  Second, instead of learning interventions to ablate a component's activations, we use corresponding activations generated from the counterfactual sentences. 

Besides these changes, training and inference with this mask proceed as in \citet{Cao_Schmid_Hupkes_Titov_2021}. At every time step, we run a forward pass of the model on an example from the \emph{Professions dataset}. We stochastically replace component outputs with corresponding counterfactual outputs, according to the mask; components with higher mask weights are replaced to a greater degree. We train the mask to induce the largest change in gendered pronoun prediction possible, while minimizing both the number of non-zero mask entries, and the magnitude of overall changes made to the model's output distribution. This procedure yields a mask over our components, whose expected values lie in $[0,1]$; higher values indicate more important components. For more details, see \Cref{app:diffmask}.

\subsection{Experiments}
We use the three methods discussed above to discover the components that cause gender bias in GPT-2 small. For CMA and DiffMask+, we limit our analysis to attention heads. All experiments were implemented using the  TransformerLens\footnote{\url{https://github.com/neelnanda-io/TransformerLens}} library \citep{nandatransformerlens2022}. For CMA, we used \citeauthor{Vig_Gehrmann_Belinkov_Qian_Nevo_Singer_Shieber_2020}'s top-$k$ strategy and selected only the top 10 heads as the NIE quickly diminishes beyond this point. Similarly, for DiffMask+, we chose the 10 heads with the highest expected mask value at the end of training. To find our circuit, we ran ACDC, finding a whole circuit containing attention heads and other components as shown in \Cref{circuit} in \Cref{sec:appendix_circuit_discovery}. For hyperparameters and training details, see \Cref{app:discovery-hyperparameters}.

\begin{figure*}[hptb]
    \centering
    \includegraphics[width=.72\linewidth,trim=35 6 30 0,clip]{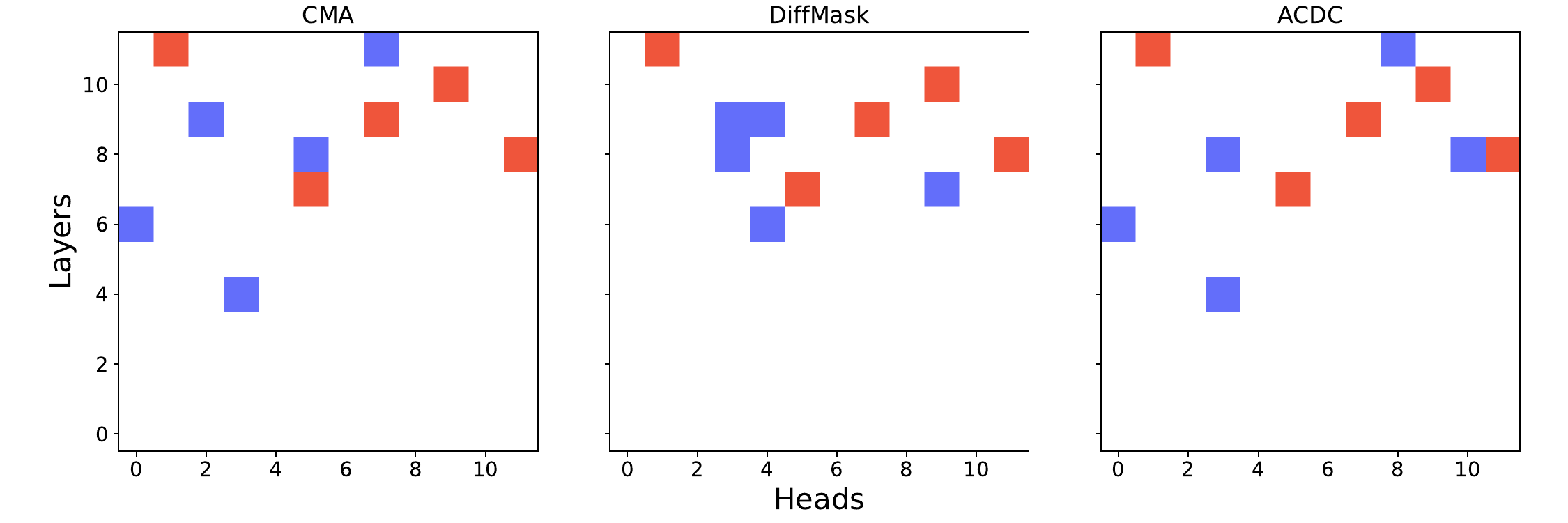}
    \includegraphics[width=0.27\linewidth,trim=3 8 3 15,clip]{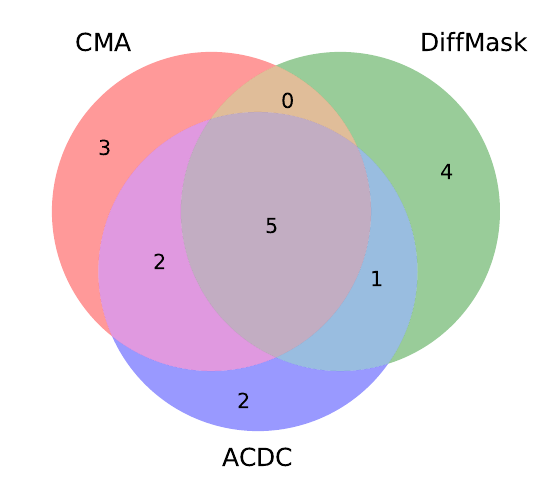}
    \caption{Top 10 attention heads selected using CMA, DiffMask+ and ACDC. Overlapping heads are shown in red.
    The Venn diagram shows the overlap counts between all combinations of the sets. }
    \label{selected_heads}
\end{figure*}

\subsection{Results}

\Cref{selected_heads} shows the attention heads selected using each method. For ACDC, we show only the attention heads from the full circuit. All methods find attention heads located mostly in the final layers of the model; this contrasts with \citet{Vig_Gehrmann_Belinkov_Qian_Nevo_Singer_Shieber_2020}, who find heads in middle layers. This may be due to the fact that \citet{Vig_Gehrmann_Belinkov_Qian_Nevo_Singer_Shieber_2020} mainly assess gender bias in co-reference resolution in their attention intervention experiments and accordingly use the WinoBias \citep{zhao-etal-2018-gender} and Winogender \citep{wino-gender} datasets. The results suggest that the dataset used for discovery influences the components picked by these methods. 

The Venn diagram in \Cref{selected_heads} shows the overlap of heads across methods. We observe a significant overlap: $5$ of the top $10$ heads are shared by all three methods. Attention heads selected using CMA and ACDC have more overlap and as observed in the mitigation results in \Cref{mitigation_results} the two methods perform similarly on different metrics. The fact that DiffMask+ yields 4 heads that are not shared might be due to its objective: DiffMask+ attempts to maximally change gendered pronoun prediction \emph{while still minimally changing the distribution overall}. This latter constraint is absent from the other two methods.

We also note that the selected heads are located in the later half of the model. We hypothesize that this may be because these heads are transferring gender information from the profession position to the end position of the sentence. Although earlier heads can also attend to gender tokens, prior work suggests that entities are enriched by lower-layer MLPs before information is extracted from them by later attention heads \citep{geva2023dissecting}.

\section{Mitigating Gender Bias}
\label{sec:intervention}
Having identified components responsible for gender bias in GPT-2 small, we test whether this information can be used to mitigate the bias.
To this end, we fine-tune the model on a dataset carefully curated to be gender balanced---this has been shown to lead to a reduction in gender bias \citep{gira-etal-2022-debiasing}.
We compare the effectiveness of fine-tuning only the components found in the previous section to various baselines, both fine-tuned and not.

\subsection{Fine-tuning Dataset and Models}

We test the effectiveness of parameter-efficient fine-tuning with the identified GPT-2 components at mitigating gender bias.
We fine-tune on the BUG dataset\footnote{\url{https://github.com/SLAB-NLP/BUG}} \cite{levy-etal-2021-collecting-large}, which contains annotated natural sentences containing one or more gendered pronouns. We use the balanced version of BUG, which has an equal number of masculine and feminine pronouns, to counteract GPT-2's gender bias in pronouns. For each model in \Cref{tab:finetuned_models}, we fine-tune only the specified subset of GPT-2's parameters and compare our methods to the not fine-tuned GPT-2 model, our \textbf{baseline}. \Cref{app:finetuning} contains fine-tuning details.

\begin{table}%
    \caption{All fine-tuned models and corresponding components selected for fine-tuning in \Cref{sec:intervention}. DM means our proposed method DiffMask+.}
    \label{tab:finetuned_models}
    \centering 
    \small
    \begin{tabular}{p{0.32\linewidth}p{0.58\linewidth}}
    \toprule
    Model Name & Selected Components\\
    \midrule  
    \textit{Full Model} &  Entire model.\\
    \textit{Random Attn Heads} & Set of 10 randomly selected attention heads \emph{not} found by CMA, ACDC or DM.\\
    \textit{All Attn Layers} &  All attention layers including the attention projection.\\
    \textit{Last 4 Attn Layers} & Last 4 attention layers.\\
    \textit{ACDC} & MLPs, attention heads,  and embedding layers found by ACDC.\\
    \textit{ACDC Attn Heads} & Attention heads from the ACDC circuit.\\
    \textit{CMA Attn Heads} & Top 10 attention heads found by CMA.\\  
    \textit{DM Attn Heads} & Top 10 attention heads found by DiffMask+.\\
    \bottomrule   
    \end{tabular}
\end{table}

\subsection{Metrics}
\label{metrics}
We use several metrics and baselines to evaluate the effectiveness of the bias mitigation under the different conditions.
To measure gender bias, we use WinoBias \citep{zhao-etal-2018-gender} and the gender bias subset of CrowS-Pairs by~\citet{neveol2022french}. We also measure model performance on the original \textit{Professions dataset} using which important components were found.
To ensure that fine-tuning did not harm models' general language modeling abilities, we also measure these, via WikiText perplexity~\citep{Merity2016PointerSM} and accuracy on BLiMP~\citep{warstadt-etal-2020-blimp-benchmark}.
All metrics, except for the perplexity, are defined as the ratio of times that the model prefers the correct/anti-stereotypical over the incorrect/stereotypical variant. Given a dataset $\mathcal D$ with pairs of stereotypical and anti-stereotypical sentences $(\mathbf{x}, \mathbf{\tilde x})$, the Stereotype Score is defined as follows. 
\begin{align}
    \text{SS} = \frac{1}{|\mathcal D|} \sum_{(\mathbf{x}, \mathbf{\tilde x}) \in \mathcal D} \mathbb I_{p(\mathbf{x}) > p(\mathbf{\tilde x})} \label{eq:ss}
\end{align}

\paragraph{WinoBias} We measure the models' gender bias using WinoBias. Even if this dataset with its small linguistic variety might not exactly reflect real-world biased language \citep{lior2023comparing}, it is widely used as its simplicity allows for controlled experiments. We measure models' gender bias using WinoBias' type 2 dataset\footnote{We choose not to discuss the results for the type 1 dataset because we do not test an actual co-reference resolution task, but rather compute the perplexities of continuing with one or the other gendered pronoun.} \citep{zhao-etal-2018-gender}.
This dataset consists of sentences containing two occupation terms and one gendered pronoun; models must determine which occupation the pronoun refers to. In type 2 examples, the sentence's syntax always determines the correct occupation (regardless of the pronoun's gender). For each sentence there is one pro- and one anti-stereotypical version, which differ only in the gender of the pronoun used. We consider a model biased if it consistently assigns higher probability to the pro-stereotypical sentence. We record the proportion of examples where the model assigns higher probability to the pro-stereotypical version.
Note that our metric differs from the original metric, which was formulated in terms of co-reference resolution accuracy. 

\paragraph{CrowS-Pairs}
The gender bias subset of CrowS-Pairs measures gender bias in LMs, construed more broadly than occupation-gender associations. It consists of minimal pairs, a more and a less stereotypical sentence. %
We consider a systematic preference for more stereotypical sentences (by comparing perplexities) to indicate a biased model. As in WinoBias, the bias is measured as the proportion of examples where the model prefers the stereotypical sentence. 
In our experiments, we use an updated version from \citet{neveol2022french} where potential validity issues (including those identified by \citet{blodgett-etal-2021-stereotyping}) have been addressed.

\paragraph{Professions} We use the \emph{Professions dataset}, with which we found bias-relevant components, to assess gender bias in the fine-tuned models. For every sentence in the dataset, we measure the probability assigned to the pro-/anti-stereotypical continuations (either \emph{he} or \emph{she}, depending on the example). We measure the proportion of examples where the pro-stereotypical continuation is more probable.

\paragraph{BLiMP}
We evaluate our models' linguistic abilities using BLiMP. 
BLiMP consists of a number of datasets, each of which targets a specific linguistic phenomenon. Each dataset contains examples, each of which is a minimal sentence pair: one sentence is correct and the other incorrect, with respect to the targeted phenomenon. 
The model should systematically assign a higher probability to the correct sentence.
We report accuracy on BLiMP as a whole, as well as on the \textit{Gender Anaphor Agreement} (AGA) and \textit{Subject Verb Agreement} (SVA) subtasks. We do this to understand the effect of our fine-tuning on these specific linguistic phenomena, where gender is only relevant for one of these tasks.

\paragraph{WikiText}
We evaluate our models' general language modeling performance by computing their perplexity on the test split of the WikiText-103 corpus\footnote{\url{https://huggingface.co/datasets/wikitext}}  (4358 examples) \citep{Merity2016PointerSM}, which consists of ``Good'' and ``Featured'' Wikipedia articles. Higher perplexity might indicate that fine-tuning hurt general language modeling abilities.

\subsection{Results}
\label{mitigation_results}

\begin{table*}
\small
\centering
\caption{Effect comparison of the different fine-tuning interventions. Reported are perplexity (PPL, measured on WikiText), three measures of linguistic adequacy (full BLiMP as well as subject-verb and anaphora agreement portions of BLiMP), and the gender bias measures from CrowS-Pairs, WinoBias, and the Professions benchmarks/datasets. The cells show the $\%$ improvement (positive is better as indicated by $\uparrow$) w.r.t. the original GPT-2 before fine-tuning, averaged over 5 seeds (absolute scores are in \Cref{sec:appendix_additional_results}). * indicates $p<0.05$ for two-sided one sample \emph{t}-test, where the original GPT-2 performance serves as the population mean.
}
\label{tab:all_results}
\begin{tabular}{llrrrrrrr}
\toprule %
 &  & \it perplexity $\uparrow$ & \multicolumn{3}{c}{\it linguistic adequacy $\uparrow$} & \multicolumn{3}{c}{\it gender bias measures $\uparrow$} \\
 &  & PPL & BLiMP & SV & AGA & CrowS. & WinoB. & Prof. \\
\midrule
\multirow[c]{2}{*}{\it baselines} & full model &  -44.2 & {\cellcolor[HTML]{D93429}} \color[HTML]{F1F1F1} -3.9* & {\cellcolor[HTML]{F47044}} \color[HTML]{F1F1F1} -2.9* & {\cellcolor[HTML]{CFEB85}} \color[HTML]{000000} 1.2* & {\cellcolor[HTML]{FECC7B}} \color[HTML]{000000} -1.4\phantom{*} & {\cellcolor[HTML]{0B7D42}} \color[HTML]{F1F1F1} 4.6* & {\cellcolor[HTML]{93D168}} \color[HTML]{000000} 2.3\phantom{*} \\
 & random attn heads & -17.0 & {\cellcolor[HTML]{F46D43}} \color[HTML]{F1F1F1} -3.0* & {\cellcolor[HTML]{FEE28F}} \color[HTML]{000000} -0.9* & {\cellcolor[HTML]{F7FCB4}} \color[HTML]{000000} 0.2\phantom{*} & {\cellcolor[HTML]{FFF8B4}} \color[HTML]{000000} -0.2\phantom{*} & {\cellcolor[HTML]{ABDB6D}} \color[HTML]{000000} 1.9\phantom{*} & {\cellcolor[HTML]{C9E881}} \color[HTML]{000000} 1.3\phantom{*} \\
\cline{1-3}
\multirow[c]{3}{*}{\it broad interventions} & all attn layers & -19.1 & {\cellcolor[HTML]{FDAF62}} \color[HTML]{000000} -2.0* & {\cellcolor[HTML]{FECA79}} \color[HTML]{000000} -1.4* & {\cellcolor[HTML]{C7E77F}} \color[HTML]{000000} 1.4* & {\cellcolor[HTML]{FEEB9D}} \color[HTML]{000000} -0.6\phantom{*} & {\cellcolor[HTML]{FAFDB8}} \color[HTML]{000000} 0.1\phantom{*} & {\cellcolor[HTML]{E8F59F}} \color[HTML]{000000} 0.6* \\
 & last 4 attn layers & -12.6 & {\cellcolor[HTML]{E65036}} \color[HTML]{F1F1F1} -3.4* & {\cellcolor[HTML]{EFF8AA}} \color[HTML]{000000} 0.4* & {\cellcolor[HTML]{FED884}} \color[HTML]{000000} -1.2* & {\cellcolor[HTML]{FFF8B4}} \color[HTML]{000000} -0.2\phantom{*} & {\cellcolor[HTML]{148E4B}} \color[HTML]{F1F1F1} 4.2* & {\cellcolor[HTML]{57B65F}} \color[HTML]{F1F1F1} 3.2* \\
 & acdc & -38.8 & {\cellcolor[HTML]{BB1526}} \color[HTML]{F1F1F1} -4.6* & {\cellcolor[HTML]{FDB567}} \color[HTML]{000000} -1.8* & {\cellcolor[HTML]{E6F59D}} \color[HTML]{000000} 0.6\phantom{*} & {\cellcolor[HTML]{FEE491}} \color[HTML]{000000} -0.9\phantom{*} & {\cellcolor[HTML]{4BB05C}} \color[HTML]{F1F1F1} 3.3* & {\cellcolor[HTML]{5DB961}} \color[HTML]{F1F1F1} 3.1\phantom{*} \\
 \cline{1-3}
\multirow[c]{3}{*}{\it narrow interventions} 
 & acdc attn heads & -16.6 & {\cellcolor[HTML]{F36B42}} \color[HTML]{F1F1F1} -3.0* & {\cellcolor[HTML]{F4FAB0}} \color[HTML]{000000} 0.3* & {\cellcolor[HTML]{F7FCB4}} \color[HTML]{000000} 0.2\phantom{*} & {\cellcolor[HTML]{42AC5A}} \color[HTML]{F1F1F1} 3.5* & {\cellcolor[HTML]{C3E67D}} \color[HTML]{000000} 1.4\phantom{*} & {\cellcolor[HTML]{ABDB6D}} \color[HTML]{000000} 1.9* \\
 & cma attn heads & -16.6 & {\cellcolor[HTML]{F36B42}} \color[HTML]{F1F1F1} -3.0* & {\cellcolor[HTML]{F5FBB2}} \color[HTML]{000000} 0.3* & {\cellcolor[HTML]{F7FCB4}} \color[HTML]{000000} 0.2\phantom{*} & {\cellcolor[HTML]{42AC5A}} \color[HTML]{F1F1F1} 3.5* & {\cellcolor[HTML]{C3E67D}} \color[HTML]{000000} 1.4\phantom{*} & {\cellcolor[HTML]{ABDB6D}} \color[HTML]{000000} 1.9* \\
 & dm attn heads & -17.5 & {\cellcolor[HTML]{F99355}} \color[HTML]{000000} -2.4* & {\cellcolor[HTML]{F8FCB6}} \color[HTML]{000000} 0.2\phantom{*} & {\cellcolor[HTML]{FFFDBC}} \color[HTML]{000000} -0.0\phantom{*} & {\cellcolor[HTML]{06733D}} \color[HTML]{F1F1F1} 4.8* & {\cellcolor[HTML]{DDF191}} \color[HTML]{000000} 0.9\phantom{*} & {\cellcolor[HTML]{6BBF64}} \color[HTML]{000000} 2.9* \\
\bottomrule
\end{tabular}
\end{table*}

\Cref{tab:all_results} presents the average bias evaluation results for CrowS-Pairs, WinoBias, and Professions, as well as for the perplexity and BLiMP metrics. 

\paragraph{Bias Metrics}
We find that all types of fine-tuning improve performance on the \textit{Professions dataset} (details in the appendix; \Cref{fig:professions}). This suggests that the fine-tuning procedure successfully changed model behavior. However, not all types of fine-tuning are equal: fine-tuning strategies that targeted late attention heads yielded models with lower stereotyping and variance than those that targeted other components, spread throughout the model.

\begin{figure}[h]
    \centering
    \includegraphics[width=0.4\textwidth]{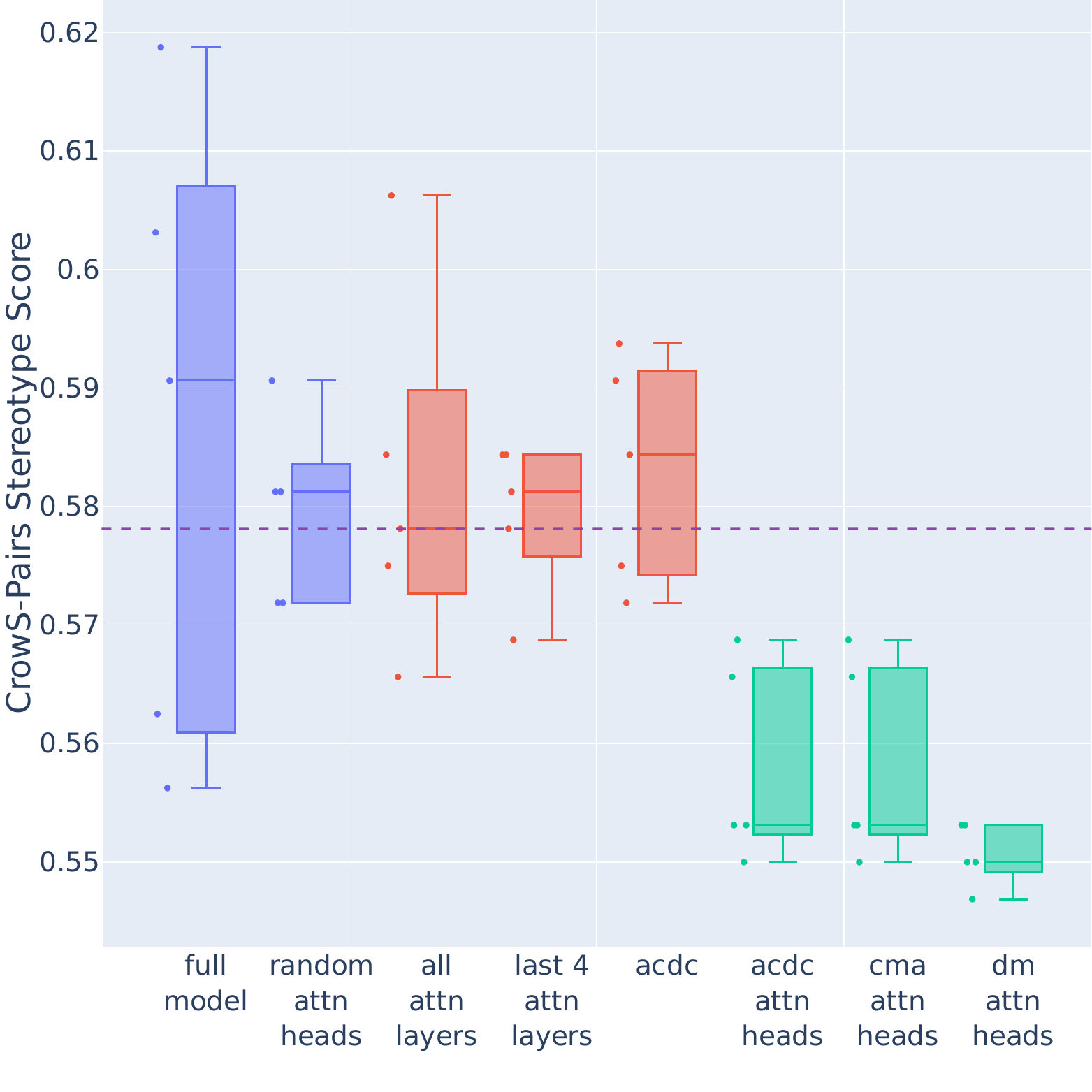}
    \caption{CrowS-Pairs results (here: lower is better). Purple models are baselines; the dotted line shows the non-fine-tuned GPT-2 performance.}
    \label{fig:crows-pairs}
\end{figure}

Similarly, the CrowS-Pairs results in \Cref{fig:crows-pairs} show that models where only the attention heads discovered using the three methods from \Cref{sec:representation-cma-circuits} were fine-tuned, achieve the best results in terms of gender bias reduction. In contrast, fine-tuning random attention heads yields no reduction in gender bias. %
The DM Attention Heads model in particular significantly reduces bias with an average stereotype score as defined in \cref{eq:ss} from $0.58$ of the baseline to $0.55$. Additionally, the scores of DM Attention Heads model have low variance while fine-tuning all attention layers, the full model, or ACDC components yields high-variance results. 

Evaluation on WinoBias yields contrasting results (\Cref{tab:all_results}). Fine-tuning the attention heads only marginally reduced the gender bias on average. Surprisingly, fine-tuning the last 4 attention layers achieved the best reduction in gender bias.
\begin{figure}[h]
    \centering
    \includegraphics[width=0.4\textwidth]{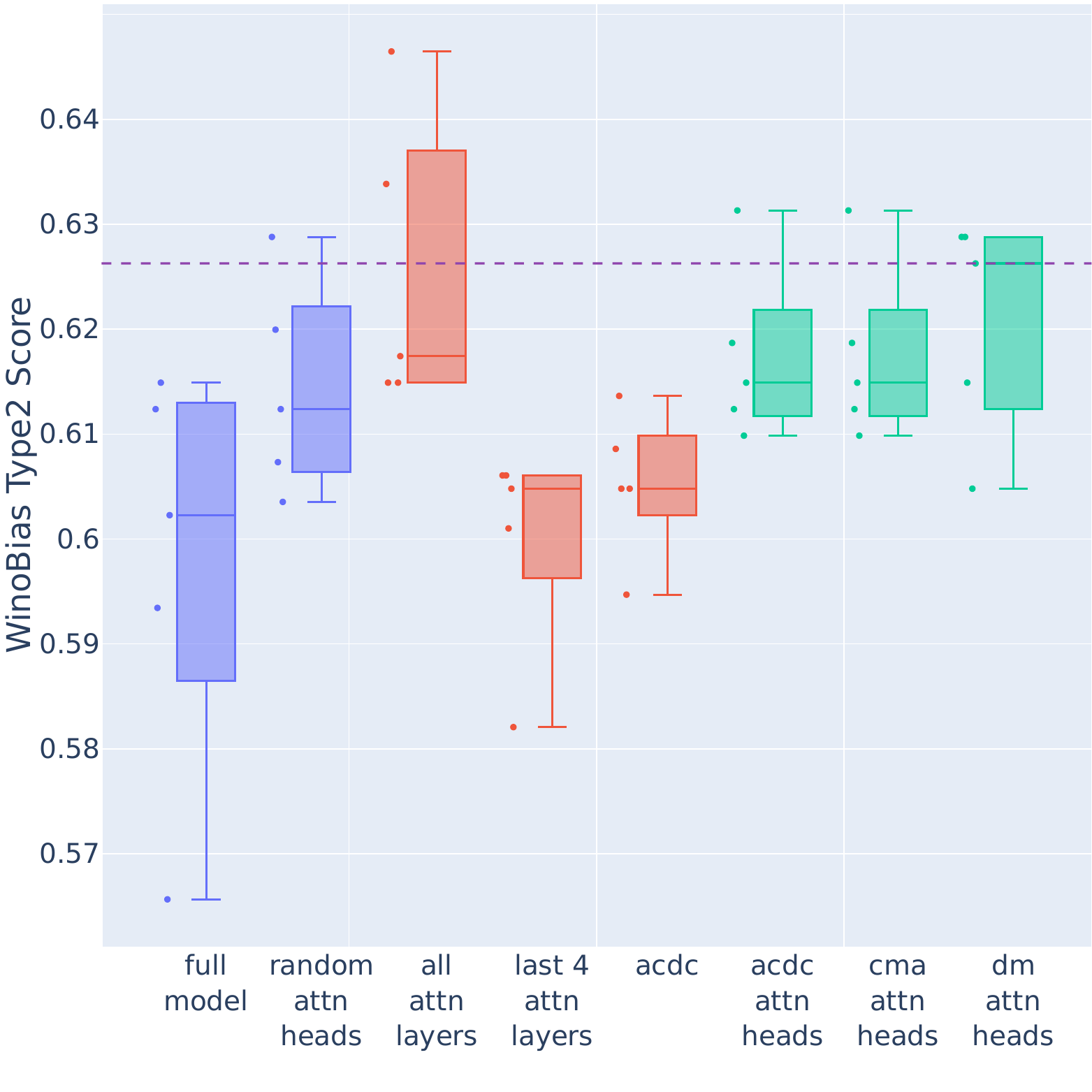}
    \caption{WinoBias Type2 Stereotype Score (here: lower is better). Purple models are baselines; the dotted line shows the non-fine-tuned GPT-2 performance.}
    \label{fig:winobias_results}
\end{figure}

At first glance, the CrowS-Pairs and WinoBias results are mixed. Fine-tuning the full model, last 4 attention layers, or ACDC components yields the most improvement on WinoBias, but these models score badly on CrowS-Pairs. However, the reverse is not true: the models that improved most on CrowS-Pairs also improved on WinoBias---although not consistently (\Cref{fig:winobias_results}). We postulate many potential explanations for the divergent outcomes seen between WinoBias and CrowS-Pairs. First, WinoBias could simply be rewarding models that perform randomly or poorly at co-reference resolution, although good overall BLiMP AGA scores suggest this is not the case. Second, gender bias in co-reference resolution might stem from a component set distinct from the ones we discovered. This is supported by \citeauthor{Vig_Gehrmann_Belinkov_Qian_Nevo_Singer_Shieber_2020}'s findings, which revealed a distinct set of attention heads that contribute to gender bias in co-reference resolution. Finally, this might be linked to how the bias measures are operationalized, which we will come back to in \Cref{sec:discussion}.

\paragraph{WikiText \& BLiMP} 
Both the perplexity measured on WikiText and accuracies on BLiMP inform us about the general language modeling capability before and after fine-tuning.
For WikiText, we observe that fine-tuning more parameters---as when we fine-tune the full model or ACDC circuit---hurts the perplexity more; the fully fine-tuned model performs the worst, increasing perplexity to 34.16 from 23.69. In contrast, targeted fine-tuning of attention heads increases perplexity by a much lower margin. This trade-off motivates finding a minimal component set to fine-tune, in order to mitigate bias while maintaining general language modeling ability.

All fine-tuned models attain lower performance on BLiMP overall than the pre-trained baseline; as in the WikiText case, the more components fine-tuned, the more performance drops. However, examining the performance on agreement subtasks reveals more nuance. On SVA, fine-tuning only the top-10 attention heads found using the methods from \Cref{sec:representation-cma-circuits} improved performance by a small margin. On AGA, almost all fine-tuned models attained scores on par with the baseline. So, while fine-tuning small sets of attention heads hurt BLiMP performance overall, the maintained performance on SVA and AGA suggest that agreement ability, gender-related or not, are not hurt.

\section{Discussion \& Conclusions}
\label{sec:discussion}

With this work, we provide an exploratory study of the identification and mitigation of gender bias in GPT-2. 
Our three different methods identify model components relevant to gender bias---according to our results, they largely agree on the most relevant attention heads: most of the heads responsible for gender bias are found mainly in the last four attention layers. 
We then intervene on each method's found components to mitigate the gender bias but maintain language modeling performance. We find that language modeling performance deteriorates only minimally for our `narrow'  interventions, but deteriorates more in conditions where a larger amount of components/parameters are adapted by fine-tuning.

Regarding computational efficiency, we find that the circuits approach is computationally inefficient compared to the other methods. For explanatory and exploratory work, like ours, circuits are very useful and can yield fine-grained insights into the model mechanisms. However, if resource efficiency is a high priority, we suggest using other methods than (automatic) circuit discovery. One key contribution of this paper is a new and very efficient method, DiffMask+, which finds a minimal set of attention heads for fine-tuning, while being computationally less prohibitive than methods such as automatic circuit discovery.

\paragraph{Limitations}
Have we reached our goal of reducing bias, using computational efficient methods?
Considering the measured gender bias, we successfully reduced the bias on two out of three datasets. This is encouraging, but our results also reveal some inconsistencies between different ways of measuring bias. 
This is not unexpected; in fact, much
previous work has highlighted many issues that put the validity and reliability of current bias measures into question~\citep[e.g.,][]{blodgett-etal-2021-stereotyping,talat2022you,dev2022measures}. Bias measures may target very different manifestations of the bias of interest~\citep{van2023undesirable}.
We therefore attribute the observed inconsistencies to the implicit versus explicit gender bias in different datasets, which could be represented differently in model components, and thus also targeted differently by fine-tuning.

Despite these challenges, we tried to address some of these concerns by using multiple different bias metrics and testing the consistency of these across different seeds. 
We believe that the success of our approach is heavily contingent upon the datasets employed for both component identification and the subsequent fine-tuning of the chosen components. For example, using template-based datasets such as WinoBias or Professions could reduce the identified components' generalizability, as components that contribute to one form of gender bias may not contribute to another. The same applies to the fine-tuning stage as well. Using a dataset with limited variability in structure might result in only partial mitigation of the behavior. We therefore conclude that for even better bias reduction, it is essential to use and develop datasets that are diverse and representative of the behaviour being studied.

\paragraph{Future work}

For a wider picture of how our findings integrate in bias identification and mitigation studies, we would like to compare our approaches to other promising methods in the literature like concept erasure at the activation level \citep[e.g., LEACE;][]{belrose2023leace} and changes to the language generation procedure \citep[e.g., ``self-debiasing'';][]{schick-etal-2021-self}.
Future work should also test whether these mitigation strategies generalize to different conditions, for example, language models larger than GPT-2 small.
Lastly, we also stress the importance of developing methodologies for operationalizing other forms of bias than binary gender in English, and to overcome difficulties we currently face when using contrastive sets and existing bias benchmarks.

\section{Acknowledgements}

OW's contributions are financed by the Dutch Research Council (NWO) as part of project 406.DI.19.059.

\newpage

\bibliography{anthology,custom,references}
\bibliographystyle{acl_natbib}

\appendix

\section{Circuit Discovery}
\label{sec:appendix_circuit_discovery}
\begin{figure}[h]
    \centering
    \includegraphics[width=\linewidth]{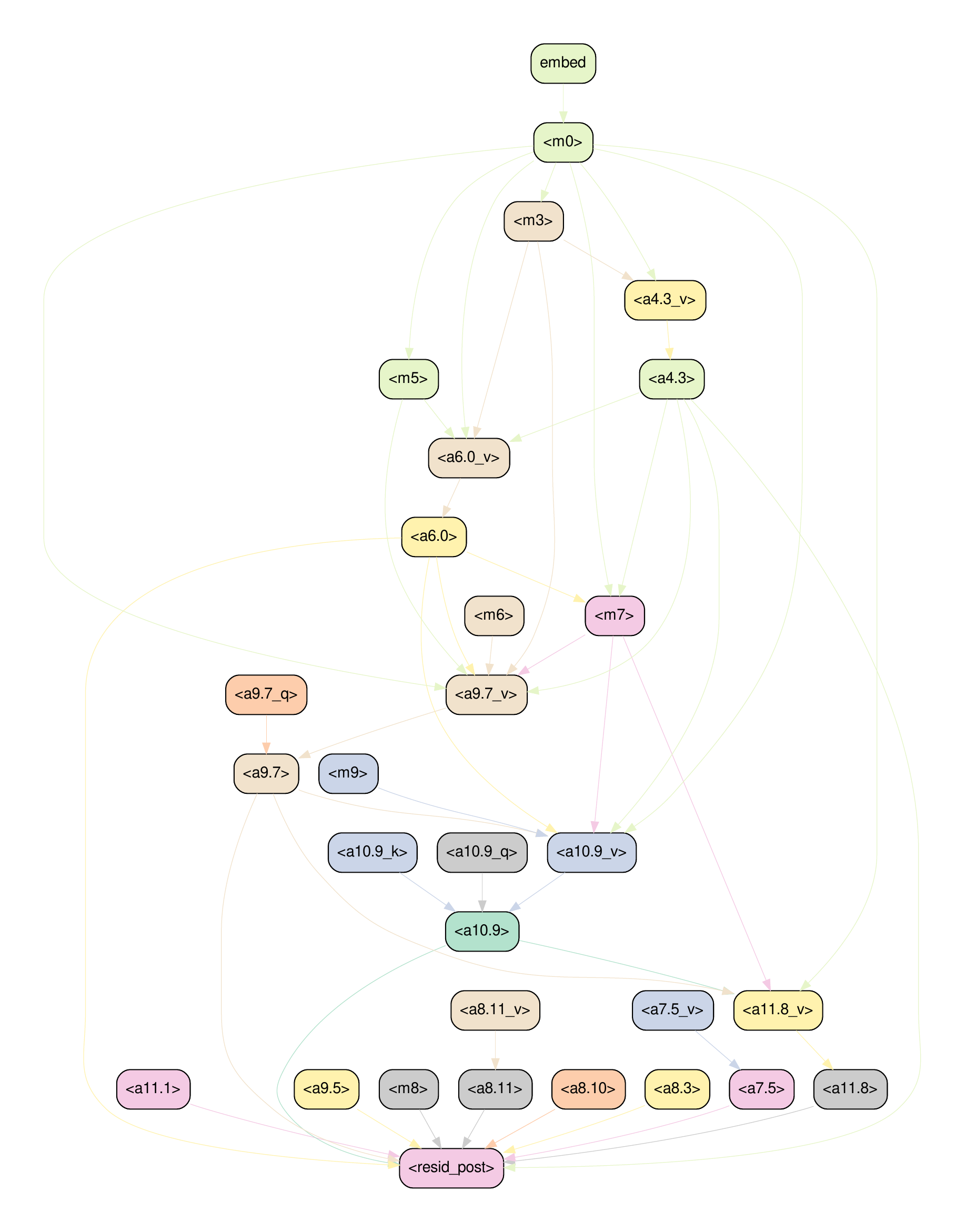}
    \caption{Circuit discovered in the GPT-2 small model using Professions dataset.}
    \label{circuit}
\end{figure}

The circuit discovered in GPT-2 small model using professions datset is shown in \Cref{circuit}

\section{DiffMask+ Implementation Details}\label{app:diffmask}
During inference, DiffMask+ works as follows. We have two inputs---our normal input $\mathbf{x}$ and our counterfactual input $\mathbf{\tilde{x}}$---as well as a $k$-dimensional binary mask $\mathbf{m}\in\{0,1\}^k$; for GPT-2 small, the number of components $k$ is $144$ as we choose to only select attention heads. We run forward passes on both inputs, recording each component's output on the normal dataset ($\mathbf{h}_1, \ldots, \mathbf{h}_k$) and the counterfactual dataset ($\mathbf{\tilde{h}}_1, \ldots, \mathbf{\tilde{h}}_k$). Finally, we run the model once more on the normal input, applying the mask: we replace each original component output $\mathbf{h}_i$ with the potentially masked output $\mathbf{h}'_i = (1-m_i) \cdot \mathbf{h}_i + m_i\cdot \mathbf{\tilde{h}}_i$\footnote{We can apply our mask either at every time step, or at only the final time step.}. If our mask captures which components are important, our masked model should behave as if it were receiving the counterfactual input.

DiffMask+'s training setup is slightly different. We cannot learn a purely binary mask, as that would not be differentiable. Instead, we learn a parameterization of a hard concrete distribution \citep{louizos2018learning}, a type of distribution that falls in $[0,1]$ and assigns non-zero probability to both 0 and 1. This distribution is parameterized by a location vector $\mathbf{z}\in[0,1]^k$, and can be sampled to produce a mask $\mathbf{m}\in [0,1]^k$. When it comes time to mask the model, we simply sample a mask from the distribution $p_{\mathbf{z}}(\mathbf{m})$; note that this mask may no longer be strictly binary. However, we can generate a deterministic and truly binary mask for use at inference time in expectation (mask set to 0 if expected value $<0.5$, and 1 otherwise).

With this setup, we can train our mask; we begin by initializing the location vector to $[0.5]^k$. We then train it on our dataset $\mathcal{D}$, optimizing a loss adapted from \citet{Cao_Schmid_Hupkes_Titov_2021} which is composed of three individual loss terms. The first, targets our task of interest---gender bias. If the original input would lead to a prediction of stereotypical pronoun $y_o$, e.g. ``she'', and corresponding anti-stereotypical pronoun is $y_c$, e.g. ``he'',  we minimize $\tilde{p}(y_o|\mathbf{x})/\tilde{p}(y_c|\mathbf{x})$ where $\tilde p$ is the intervened or masked model's output distribution. This is minimized when the anti-stereotypical prediction is much more likely than the original stereotypical prediction, i.e. when the relevant model components are intervened with the corresponding counterfactual output. 

The second loss term is the expected number of non-zero elements in our sampled mask; we want our mask to be sparse. Ideally, this would be a hard constraint, where the number of non-zero elements is $\leq\alpha$ for a chosen $\alpha$; we will instead use a Lagrangian relaxation of this constraint. The third term is the KL divergence between the unmasked model's output distribution $p(y|\mathbf{x})$ and masked model's output distribution $\tilde{p}(y|\mathbf{x})$ ; we want our masking to minimally change model output, besides task-relevant output. Formally, and much like \citet{Cao_Schmid_Hupkes_Titov_2021}, %
we optimize:
\begin{equation}
    \begin{split}
    \max_\lambda\min_{\mathbf{z}} &\sum_{\mathbf{x}, y_o, y_c\in\mathcal{D}} \frac{\tilde{p}(y_o|\mathbf{x})}{\tilde{p}(y_c|\mathbf{x})}\\ &+\lambda\left(\sum_{i=1}^{k}\mathbb{E}_{p_{z_i}(m_i)}[m_i\neq 0] - \alpha\right)\\ 
    &+ \beta D_{KL}(p(y|\mathbf{x}) || \tilde{p}(y|\mathbf{x}))
    \end{split}
\end{equation}

Here, $\alpha$ and $\beta$ are hyperparameters regulating sparsity and KL-divergence weight, respectively; $\lambda\in\mathbb{R}_{\geq 0}$ is our Lagrangian multiplier. Optimizing this loss should produce a mask that captures the components relevant to gender bias, while being maximally sparse, and still mostly preserving the model's output distribution.

\section{Component Discovery Hyperparameters}\label{app:discovery-hyperparameters}
We optimized the DiffMask loss using Adam \citep{Kingma2014AdamAM} for $200$ epochs on the professions dataset with a learning rate $10^{-3}$ and a constant schedule. We choose the sparsity hyperparameter $\alpha = 10$ for selecting 10 attention heads and the KL-Divergence weight $\beta = 1$ as proposed in \citet{Cao_Schmid_Hupkes_Titov_2021}.  At the end of the training, we choose the top-10 heads with the highest expected value of the location parameter of the stochastic mask. 

For the ACDC experiment, we chose a threshold of 0.01, eliminating edges if ablating them caused a change in performance of less than 0.01, as measured by our pronoun probability difference metric.

\section{Fine-tuning experiment}
\label{app:finetuning}
In \Cref{sec:intervention}, we fine-tune each model for a maximum of $20$ epochs using AdamW optimizer \citep{Loshchilov2017FixingWD} with an initial learning rate $10^{-4}$ and a linear schedule. We optimize Cross Entropy Loss. The BUG balanced dataset contains $25844$ sentences, which we split into gender-balanced training and validation sets, containing 90\% and 10\% of the data respectively.  We use the validation loss both for selecting the best model and early stopping with a patience of $10$ epochs.

\begin{table*}[!h]
\small
\centering
\caption{Comparison of the effect of the different fine-tuning interventions. Reported are perplexity (PPL, measured on WikiText), three measures of linguistic adequacy (full BLiMP, and subject-verb and anaphora agreement portions of BLiMP), as well as the gender biases measures from CrowS-Pairs, WinoBias, and the Professions benchmarks/datasets.
}
\label{tab:absolute_results}
\begin{tabular}{llrrrrrrr}
\toprule %
 &  & \it perplexity & \multicolumn{3}{c}{\it linguistic adequacy} & \multicolumn{3}{c}{\it gender bias measures} \\
 &  & PPL & BLiMP & SV & AGA & CrowS. & WinoB. & Prof. \\
\midrule
\multirow[c]{2}{*}{baseline} & original gpt2 & 23.69 & 0.80 & 0.90 & 0.95 & 0.58 & 0.63 & 0.84 \\
& full model & 34.16 & 0.77 & 0.87 & 0.97 & 0.59 & 0.60 & 0.82 \\
& random attn heads & 27.72 & 0.77 & 0.89 & 0.96 & 0.58 & 0.61 & 0.83 \\
\cline{1-9}
\multirow[c]{3}{*}{\it broad interventions} & all attn layers & 28.22 & 0.78 & 0.89 & 0.97 & 0.58 & 0.63 & 0.83 \\
 & last 4 attn layers & 26.67 & 0.77 & 0.90 & 0.94 & 0.58 & 0.60 & 0.81 \\
 & acdc & 32.89 & 0.76 & 0.88 & 0.96 & 0.58 & 0.61 & 0.81 \\
 \cline{1-9}
\multirow[c]{3}{*}{\it narrow interventions} & acdc attn heads & 27.62 & 0.77 & 0.90 & 0.96 & 0.56 & 0.62 & 0.82 \\
 & cma attn heads & 27.62 & 0.77 & 0.90 & 0.96 & 0.56 & 0.62 & 0.82 \\
 & dm attn heads & 27.84 & 0.78 & 0.90 & 0.95 & 0.55 & 0.62 & 0.81 \\
\bottomrule
\end{tabular}
\end{table*}

\section{Additional Results}
\label{sec:appendix_additional_results}
\begin{figure}[h]
    \centering
    \includegraphics[width=1.0\linewidth]{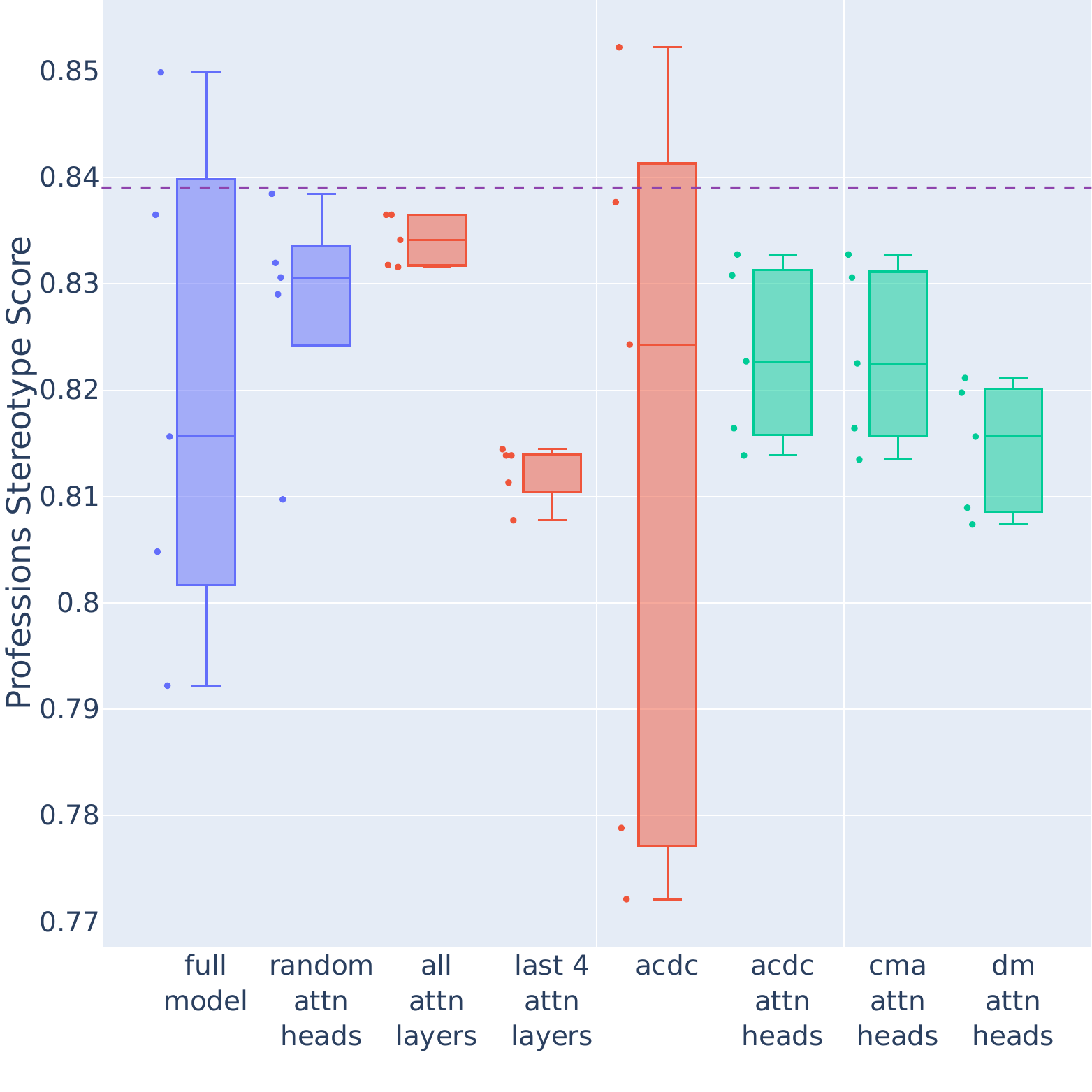}
    \caption{Professions Stereotype Score (here: lower is better). Purple models are baselines; the dotted line shows the non-fine-tuned GPT-2 performance.}
    \label{fig:professions}
\end{figure}

\begin{figure}[h]
    \centering
    \includegraphics[width=1.0\linewidth]{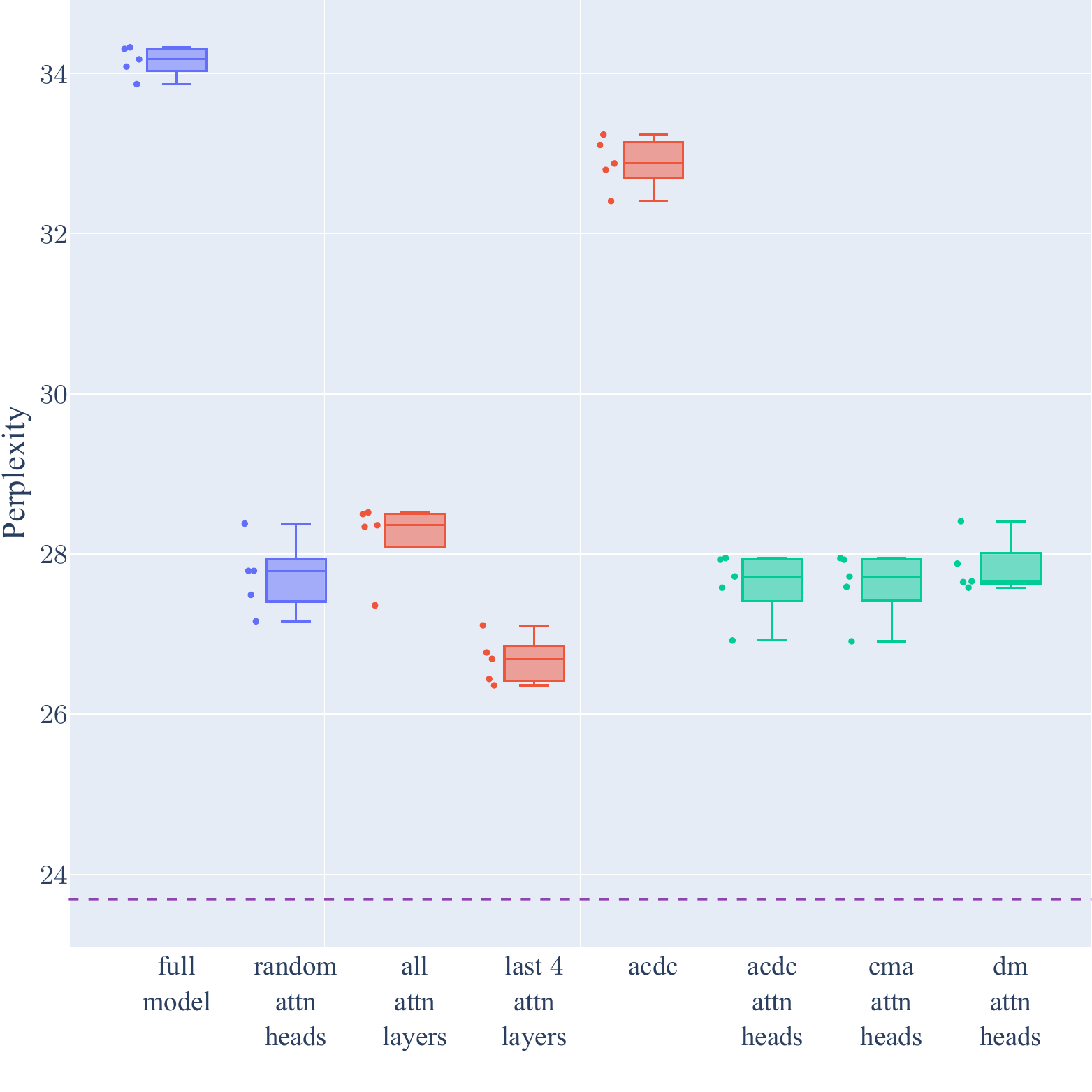}
    \caption{Test perplexity (lower is better) on WikiText-103. Purple models are baselines; the dotted line shows the non-fine-tuned GPT-2 performance.}
    \label{wikitext-103-perplexity}
\end{figure}

\begin{figure}[h]
    \centering
    \includegraphics[width=1.0\linewidth]{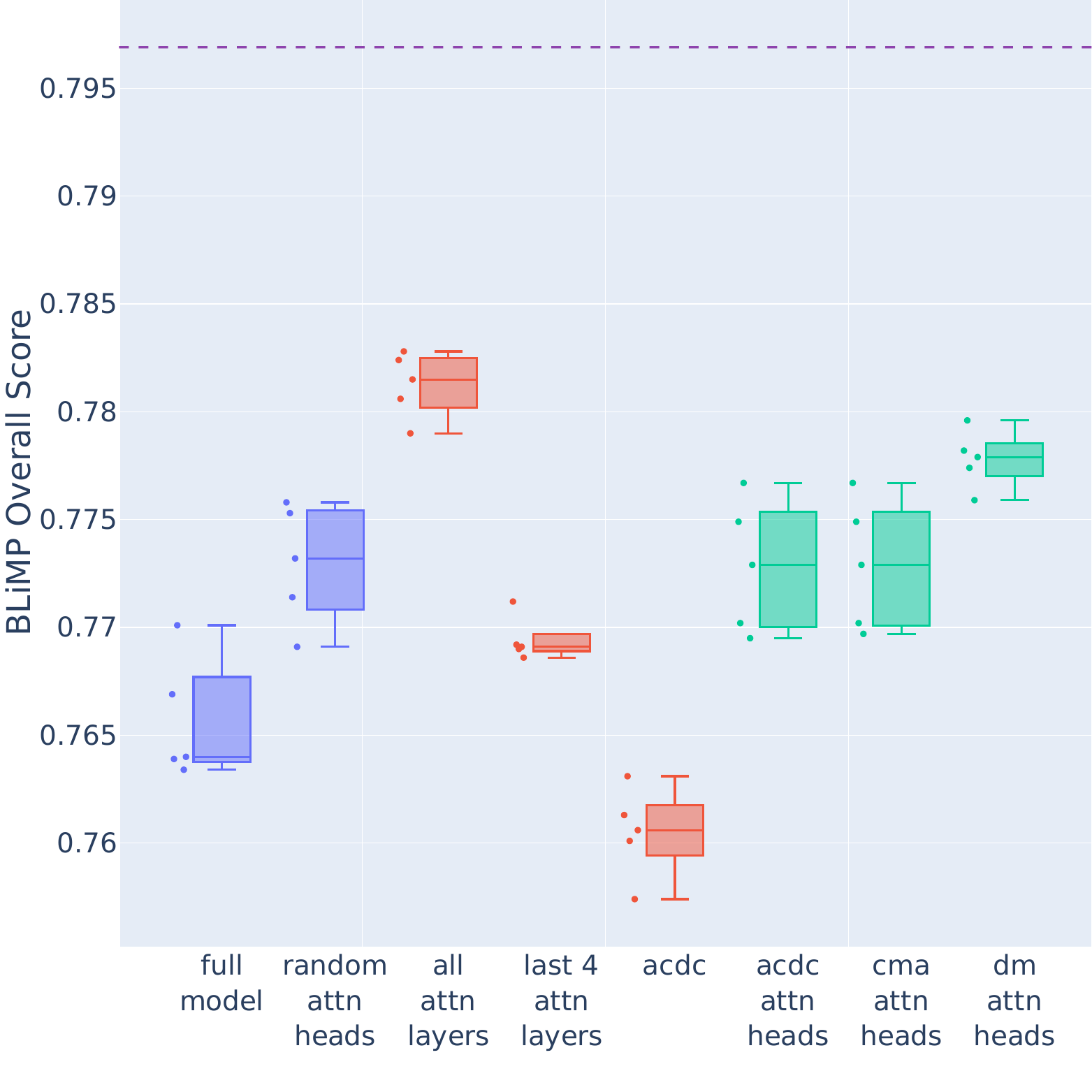}
    \caption{BLiMP Overall results (higher is better). Purple models are baselines; the dotted line shows the non-fine-tuned GPT-2 performance.}
    \label{blimp_overall}
\end{figure}

\begin{figure}[h]
    \centering
    \includegraphics[width=1.0\linewidth]{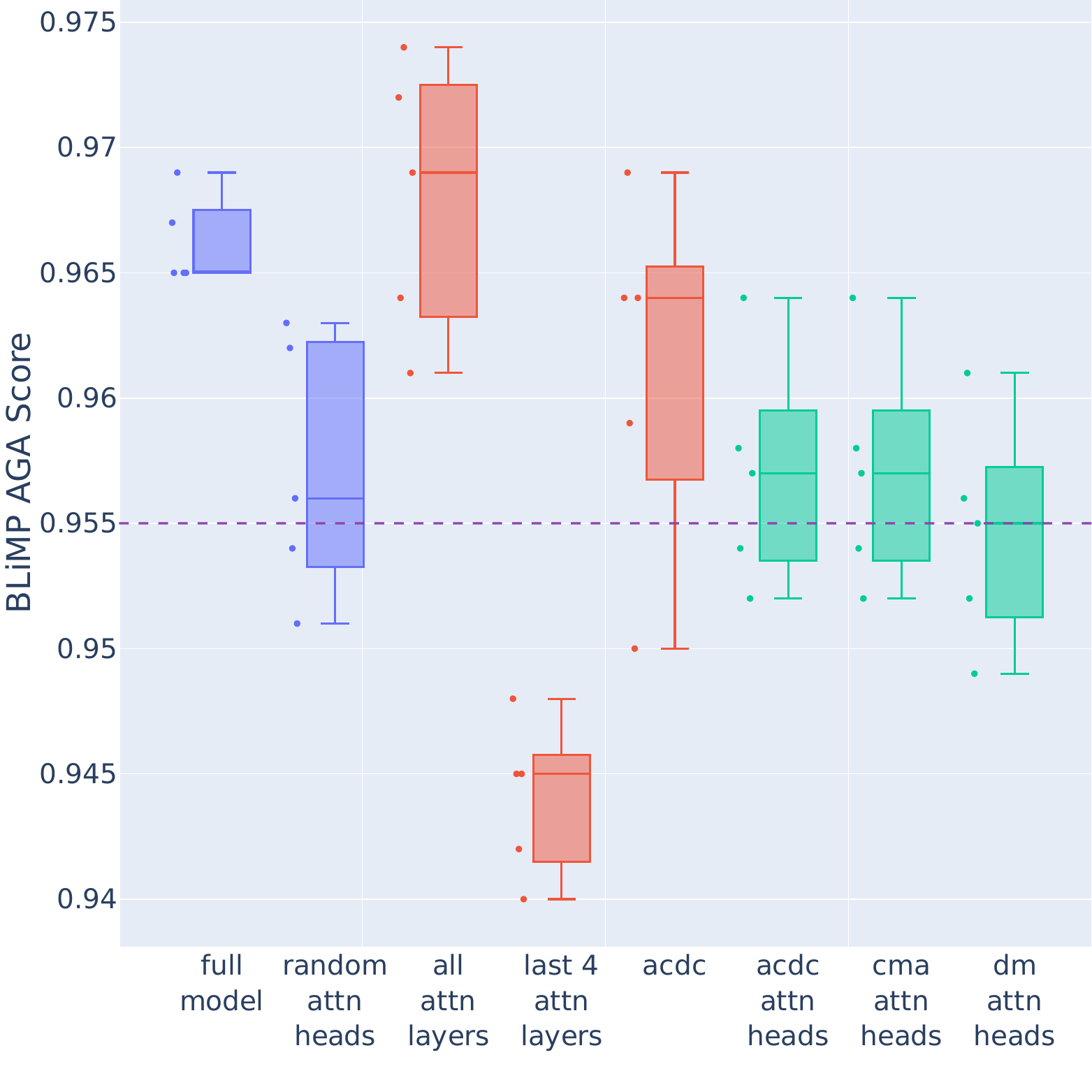}
    \caption{BLiMP Anaphor Gender Agreement results (higher is better). Purple models are baselines; the dotted line shows the non-fine-tuned GPT-2 performance.}
    \label{blimp_aga}
\end{figure}

\begin{figure}[h]
    \centering
    \includegraphics[width=1.0\linewidth]{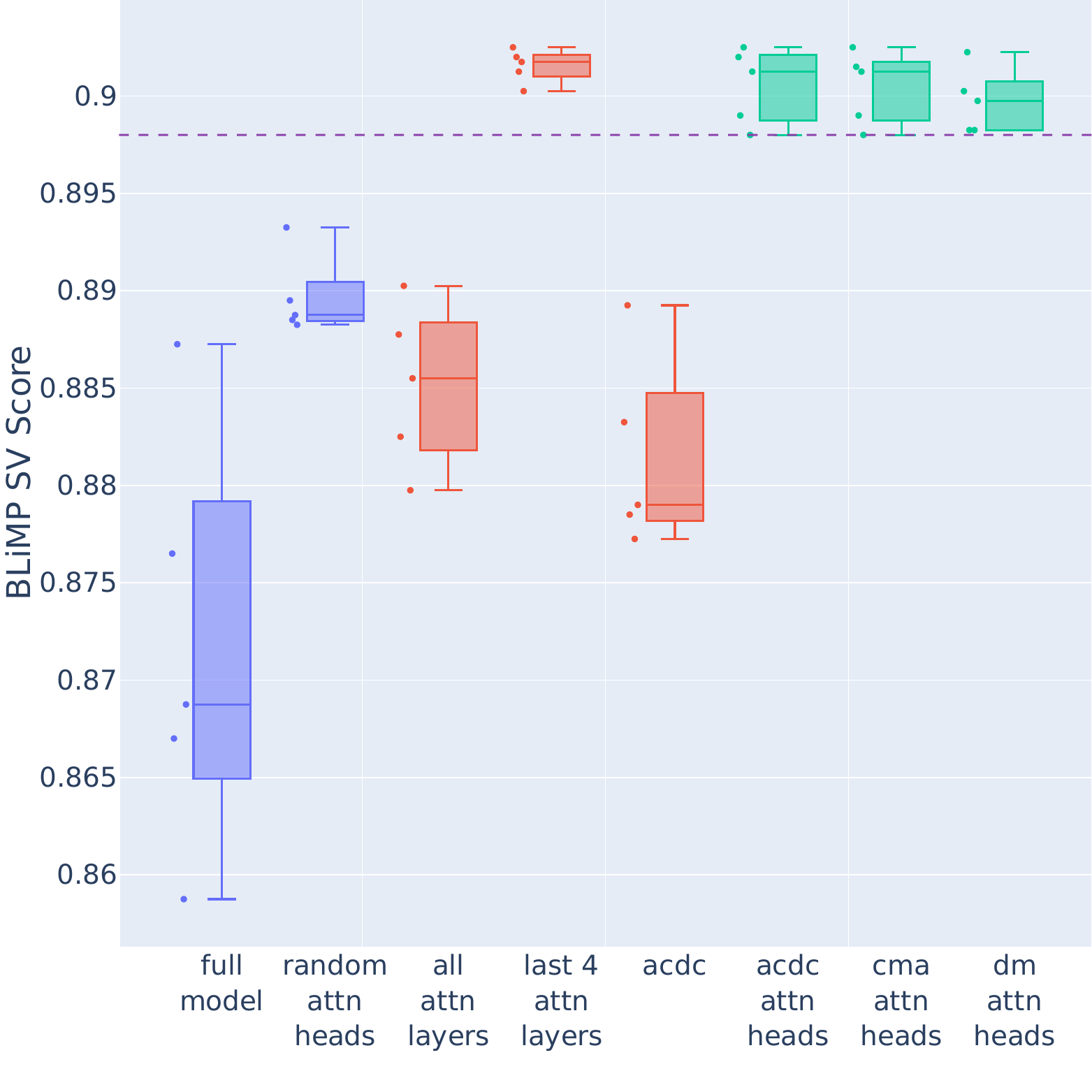}
    \caption{BLiMP Subject Verb Agreement results (higher is better). Purple models are baselines; the dotted line shows the non-fine-tuned GPT-2 performance.}
    \label{blimp_sv}
\end{figure}

 \Cref{tab:absolute_results} shows all results of fine-tuned models and baselines rounded to up to $2$ decimals. 
 \Cref{fig:professions} shows the stereotype scores of different models evaluated on the Professions dataset.  \Cref{wikitext-103-perplexity} shows the perplexity of different models evaluated on WikiText-103. 
\Cref{blimp_overall} shows the BLiMP overall results measured over 5 different iterations. Similarly, \Cref{blimp_aga} and \Cref{blimp_sv} shows the AGA and SVA results respectively.

\end{document}